%% file: main.tex
\documentclass[10pt,twocolumn,letterpaper]{article}

\usepackage[pagenumbers]{cvpr} 
\usepackage[accsupp]{axessibility} 
\usepackage{array}
\newcolumntype{C}{>{\centering\arraybackslash}X}
\newcolumntype{P}[1]{>{\centering\arraybackslash}p{#1}}
\newcolumntype{L}[1]{>{\raggedright\arraybackslash}p{#1}}  
\usepackage{multirow}
\usepackage{placeins}


\definecolor{cvprblue}{rgb}{0.21,0.49,0.74}
\usepackage[pagebackref,breaklinks,colorlinks,allcolors=cvprblue]{hyperref}

\def\paperID{7} 
\def\confName{CVPR}
\def\confYear{2025}

\title{SRVP: Strong Recollection Video Prediction Model Using Attention-Based Spatiotemporal Correlation Fusion}

\author{
    Yuseon Kim$^{1,2}$ \quad
    Kyongseok Park$^{1,2}$\thanks{Corresponding author} \\
    $^1$Korea Institute of Science and Technology Information (KISTI) \\
    $^2$Department of Applied AI, University of Science and Technology (UST) \\
    {\tt\small {\{yskblue,gspark\}@kisti.re.kr}}
}

\begin{document}
\maketitle
\input{sec/0_abstract}    
\input{sec/1_intro}
\input{sec/2_background}
\input{sec/3_methods}
\input{sec/4_experiments}
\input{sec/5_conclusions}
\input{sec/ack}
{
    \small
    \bibliographystyle{ieeenat_fullname}
    \bibliography{main}
}


\end{document}

%% file: sec/0_abstract.tex
\begin{abstract}
Video prediction (VP) generates future frames by leveraging spatial representations and temporal context from past frames. Traditional recurrent neural network (RNN)-based models enhance memory cell structures to capture spatiotemporal states over extended durations but suffer from gradual loss of object appearance details. To address this issue, we propose the strong recollection VP (SRVP) model, which integrates standard attention (SA) and reinforced feature attention (RFA) modules. Both modules employ scaled dot-product attention to extract temporal context and spatial correlations, which are then fused to enhance spatiotemporal representations. Experiments on three benchmark datasets demonstrate that SRVP mitigates image quality degradation in RNN-based models while achieving predictive performance comparable to RNN-free architectures.
\end{abstract}

%% file: sec/1_intro.tex
\section{Introduction}
\label{sec:intro}

Video prediction (VP) involves generating future video sequences by learning spatial correlations and temporal information from past frames; it is also considered a spatiotemporal forecasting problem. Unlike single-image prediction, VP requires estimating temporal dynamics and spatial representations simultaneously, making it a complex and challenging task.
 
Traditional approaches for processing image-sequence datasets, such as videos, rely on recurrent neural network (RNN)-based models such as long short-term memory (LSTM) networks \cite{lstm, lstm2}, which excel in time series forecasting and have applications beyond VP. 
Early VP models used a simple LSTM-based encoder-decoder framework \cite{vplstm}, where the encoder compresses input into a lower-dimensional hidden representation, and the decoder predicts future frames. Subsequently, convolutional LSTM (ConvLSTM) \cite{convlstm} extended LSTM’s input-to-state and state-to-state transitions into a convolutional structure.

Building on ConvLSTM, various VP models have been proposed to enhance long-term video dependency capture \cite{predrnn, e3dlstm, predrnn2, mim, conttlstm, phydnet, saconvlstm, motionrnn, stam}.
These methods include augmenting ConvLSTM cells with spatiotemporal states and deepening state circulation through zigzag transitions  \cite{predrnn}, extending dual-memory units into a cascaded structure along the time axis \cite{predrnn2}, modeling non-stationary and stationary spatiotemporal dynamics \cite{mim}, generalizing ConvLSTM to higher dimensions \cite{conttlstm}, integrating transient and motion trends in video sequences \cite{motionrnn}, and incorporating self-attention into ConvLSTM cells \cite{saconvlstm}.

Although these RNN-based methods can naturally model the temporal transitions of video, they can be highly sensitive to object motion. Spatiotemporal memory cells store information from previous time steps and update the spatial information of the next time step accordingly. Owing to this inherent characteristic, the model tends to predict average values in regions of change, minimizing error but leading to blurring in generated frames. Moreover, errors accumulate over time, further degrading accuracy in long-term predictions.

In addition, vision transformer (ViT)-based methods have been explored for video analysis \cite{odtrans, hrformer, swintrans, convtrans, attvit}. However, most focus on extracting high-dimensional image features for tasks such as classification, with limited attempts at VP. ViT-based VP approaches \cite{vpvit, mimo} aim to mitigate error accumulation in long-term predictions by generating future frames in a single step. However, unlike RNN models that employ gating mechanisms to regulate temporal variations, ViTs lack an explicit mechanism for controlling temporal changes. Therefore, additional techniques such as positional encoding and 3D convolutions are required for auto-regression, but their effectiveness in capturing both short- and long-term dependencies remains unclear.

Moreover, input image size significantly impacts ViT-based VP performance. As ViTs partition images into non-overlapping patches, larger images force each patch to cover a broader area, reducing fine detail capture and potentially decreasing accuracy. In contrast, smaller patches increase the number of boundaries within an image, disrupting spatial continuity. In addition, a larger number of patches also raises computational costs and may require extensive hyperparameter tuning to identify the optimal patch size.

Therefore, we propose the strong recollection VP (SRVP) model, which integrates an attention-based approach to improve the retention of spatially varying object representations over longer durations while preserving the ability of RNNs to estimate temporal context. SRVP enhances long-term memory of moving objects by referencing spatial features from past frames through temporal attention and capturing spatial correlations between pixels via spatial self-attention.

%% file: sec/2_background.tex
\section{Background}
\label{sec:background}

The ConvLSTM unit enables spatiotemporal modeling through three interacting gates: input, forget, and output \cite{conttlstm}. However, because ConvLSTM merely extends conventional LSTM gate operations into a convolutional structure, it inherently prioritizes temporal variations.

To address this limitation and enhance the extraction of spatiotemporal representations, spatiotemporal LSTM (ST-LSTM) \cite{predrnn} was introduced. The ST-LSTM unit employs a dual-memory structure, integrating a spatiotemporal memory cell alongside the conventional ConvLSTM memory cells. This design allows the model to capture both low-dimensional image features and high-level abstract representations by updating spatiotemporal memory cell states in a zigzag pattern. However, this complex nonlinear transition structure is prone to the vanishing gradient problem, where gradient magnitudes decay exponentially during backpropagation through time (BPTT). To mitigate this, causal LSTM and the gradient highway unit (GHU) were proposed \cite{predrnn2}. Causal LSTM extends the dual-memory structure of ST-LSTM into a cascaded architecture, improving structured transitions across deeper temporal stages, expanding the receptive field, and enhancing long-term retention of object features. In addition, the GHU shortens backpropagation paths from distant inputs to future outputs, reducing gradient decay.  Subsequently, the memory in memory (MIM) \cite{mim} block was introduced to capture complex spatiotemporal variations. Using a two-cascaded, self-renewing memory module, MIM leverages difference signals between adjacent recurrent states to capture both stationary and non-stationary spatiotemporal dynamics. A network with stacked MIM blocks enabled the modeling of high-order non-stationary phenomena in complex systems such as traffic flow and weather patterns.

The gating mechanism in these RNN-based approaches sophisticatedly regulates information retention and forgetting, allowing for the effective learning of natural temporal information. These models update the spatiotemporal state for the next frame based on information from prior time steps. However, in regions with continuous motion---for instance, with a walking pedestrian or a moving vehicle---repeated updates degrade spatial information retention, weakening object appearance recall over time.

Consequently, the model tends to predict average values in dynamically changing areas to minimize error, leading to more pronounced blurring effects in motion regions compared to static ones. As VP tasks require not only high quantitative performance but also visually coherent frame generation, reducing blurring artifacts remains a critical challenge.

Accurately preserving the intrinsic representation of moving objects at each time step is key to generating sharper and more natural video sequences. The proposed method follows this principle by introducing two attention modules designed to preserve spatially varying object representations while incorporating the temporal context retention capability of RNNs. The subsequent section details our approach.

%% file: sec/3_methods.tex
\section{Methods}
\label{sec:methods}

\begin{figure}[t]
  \centering
   \includegraphics[width=\columnwidth]{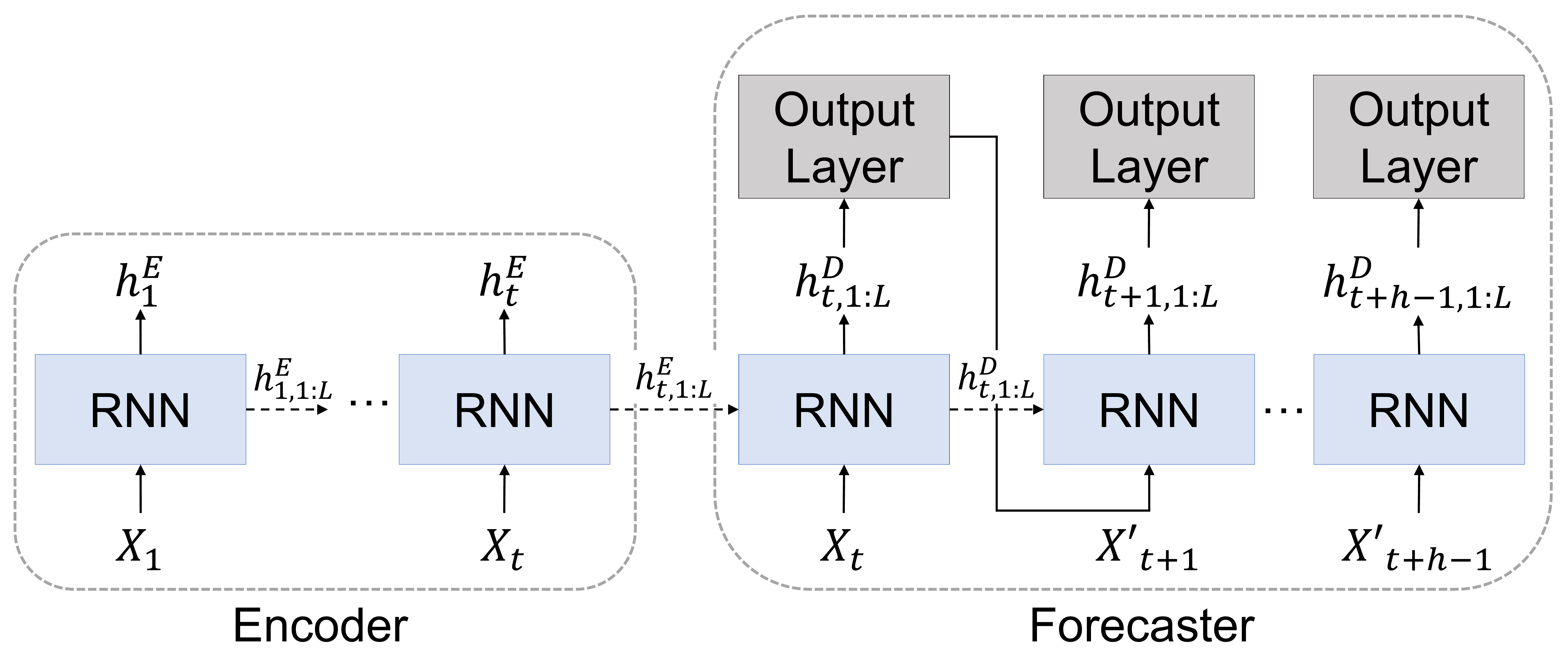}
   \caption{Baseline VP model. The encoder network includes $L$-RNN layers, whereas the forecaster network comprises $L$-RNN layers and a CNN-based output layer.}
   \label{fig1}
\end{figure}
\subsection{Baseline Architecture}
\label{sec:base}
We adopted an RNN-based encoder-forecaster architecture \cite{s2s, gru} as the baseline VP model, as depicted in \cref{fig1}. This architecture comprises encoding and forecasting components, resembling the ConvLSTM encoder-decoder model \cite{convlstm}. The encoding network comprises $L$-RNN layers, whereas the forecasting network comprises $L$-RNN layers and a convolutional neural network (CNN)-based output layer. The encoding network sequentially processes $N$ historical frames, $X_{1:t} \in \mathbb{R}^{N \times C \times H \times W}$, as input, learning the video dynamics over time. Each image frame is encoded layer by layer, progressively transforming it into $h_{1:t}^{E} \in \mathbb{R}^{N \times M \times H \times W}$, which captures high-dimensional spatiotemporal features. The hidden states of each layer are propagated along the temporal axis (horizontally), allowing for simultaneous updates of both temporal context and spatial representations. In the forecasting network, the final hidden state of the encoder, $h_{t, 1:L}^{E} \in \mathbb{R}^{L \times M \times H \times W}$, serves as the initial state for predicting future frame sequences. 

In this study, the number of layers ($L$) was four, and the size of the hidden states ($M$) in each layer was 128 in both the encoder and the forecaster.

\subsection{Convolutional Gated Recurrent Unit}
\label{sec:convgru}
This study used the convolutional gated recurrent unit (ConvGRU) as the RNN layer of SRVP. Similar to ConvLSTM, ConvGRU is a spatiotemporal recurrent unit that extends the conventional GRU \cite{gru, gru2} by incorporating spatial dimensions. It comprises two principal gating mechanisms: reset gate $r_{t}$ and update gate $z_{t}$, which are formulated as follows:
\begin{equation}\label{eq1}
    \begin{split}
    &r_{t} = \sigma(W_{xr} \ast X_{t} + W_{hr} \ast h_{t-1} + b_{r}) \\
    &g_{t} = \tanh(W_{xg} \ast X_{t} + W_{hg} \ast (r_{t} \circ h_{t-1}) + b_{g}) \\
    &z_{t} = \sigma(W_{xz} \ast X_{t} + W_{hz} \ast h_{t-1} + b_{z}) \\
    &h_{t} = (1-z_{t}) \circ g_{t} + z_{t} \circ h_{t-1}
    \end{split}
\end{equation}
where $\sigma$ represents the sigmoid function, and $\ast$ and $\circ$ denote the convolution operator and Hadamard product, respectively. The reset gate determines how much of the past information should be forgotten by considering both the current input and historical data, while the update gate regulates the extent to which newly computed information, along with past information, is incorporated into the current state.

The ConvGRU unit learns spatiotemporal representations through a more simplified hidden state computation than ConvLSTM. This streamlined architecture improves the responsiveness and efficiency of the model in capturing temporal variations \cite{gru2}.

\begin{figure}[t]
  \centering
   \includegraphics[width=0.85\columnwidth]{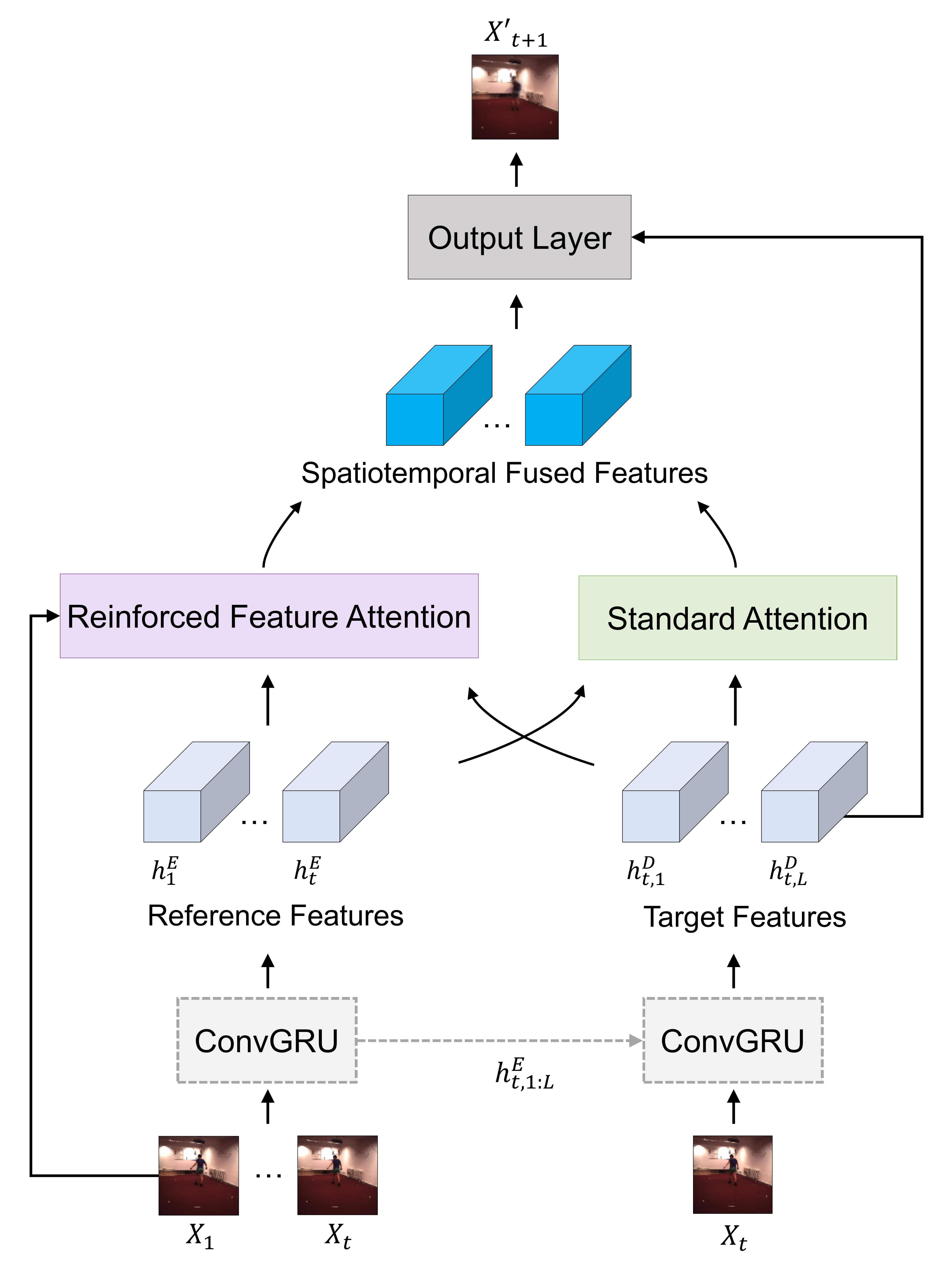}
   \caption{Overall SRVP architecture, primarily comprising two modules, standard attention (SA) and reinforced feature attention (RFA), which together extract spatiotemporal fused features.}
   \label{fig2}
\end{figure}

\cref{fig2} illustrates the overall architecture of the SRVP model. When the prediction target is $X_{t+1}$, the ConvGRU layers in the encoder take the historical frames, $X_{1:t}$, as input and generate reference features $h_{1:t}^{E}$. The ConvGRU layers in the forecaster receive the current frame, $X_{t}$, and the final hidden state of the encoder as initial states, producing the target features $h_{t, 1:L}^{D} \in \mathbb{R}^{L \times M \times H \times W}$.

\subsection{Spatiotemporal Correlation Fusion}
\label{sec:srvp}
To effectively predict both short- and long-term variations in video sequences, a spatiotemporal prediction model must fulfill the following two key requirements:
\vspace{\baselineskip}
\begin{enumerate}
  \item The spatiotemporal memory should concentrate on capturing the temporal context of the video (e.g., the position of objects in the next frame) by appropriately storing and updating information.
  \item The model should accurately recall the intrinsic representation of objects (e.g., the structure of a human body or the shape of digits).
\end{enumerate}
\vspace{\baselineskip}
To address this challenging task, we propose a method that leverages the attention mechanism to extract various spatiotemporal features from video sequences. The proposed approach initially focuses separately on temporal context estimation and spatial correlation enhancement and then integrates these two distinct feature representations to generate spatiotemporal features, which are used to generate subsequent frames.

As shown in \cref{fig2}, SRVP extends the baseline architecture by incorporating attention modules. It comprises two main modules, standard attention (SA) and reinforced feature attention (RFA), which together extract integrated spatiotemporal features.

\begin{figure}[t]
  \centering
   \includegraphics[width=0.85\columnwidth]{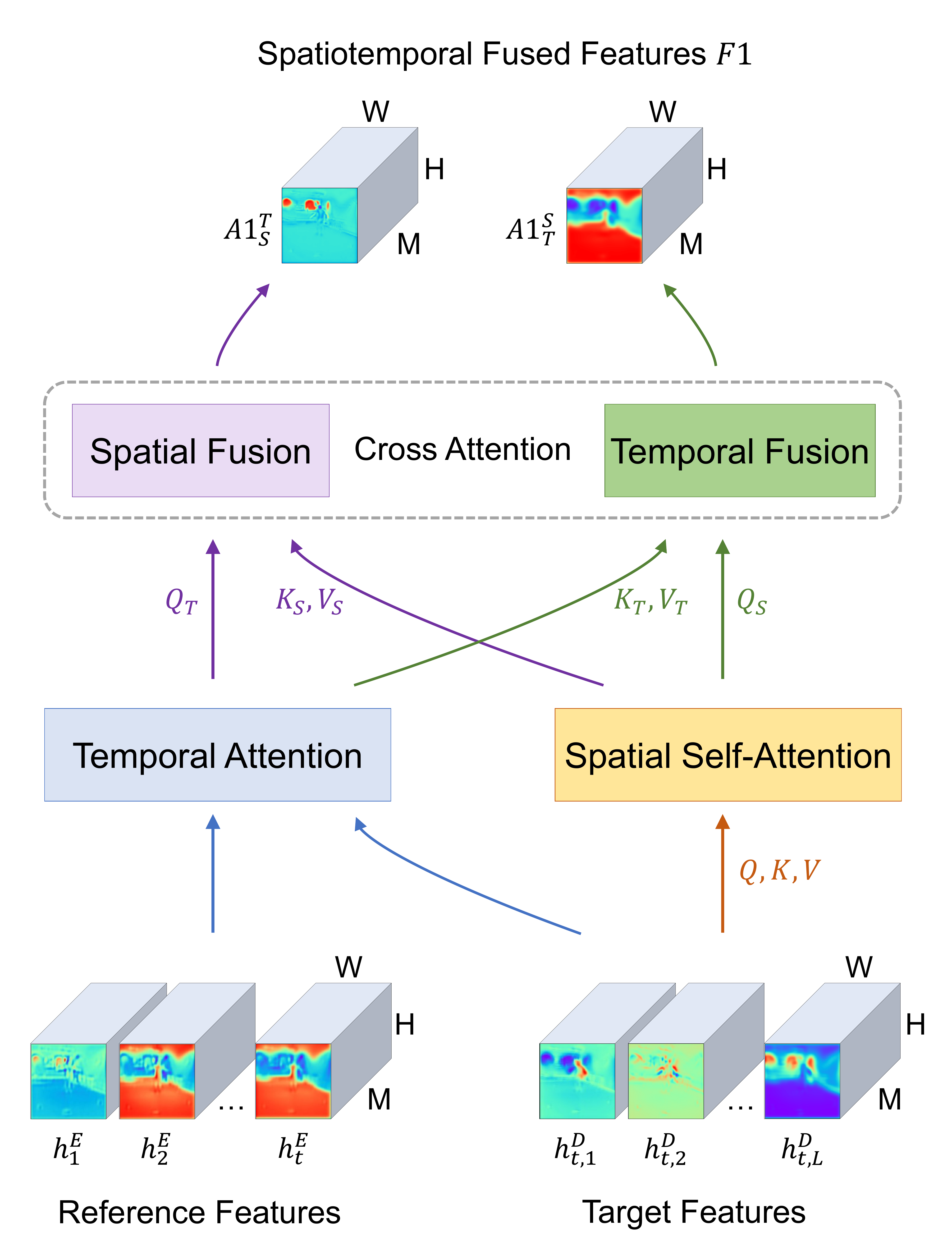}
   \caption{SA module of SRVP. This module performs temporal attention, spatial self-attention, and cross-attention.}
   \label{fig3}
\end{figure}

The SA module comprises temporal attention, spatial self-attention, and cross-attention (\cref{fig3}). In machine translation, the attention mechanism identifies the most critical elements within an input sentence that contribute to understanding the overall context \cite{att}. From the VP perspective, this can be analogous to identifying the most relevant frames from the historical sequences that best inform the current prediction. Furthermore, the self-attention mechanism \cite{selfatt} is formulated as a process for exploring spatial correlations between pixels in the spatiotemporal features.

\begin{figure}[t]
  \centering
   \includegraphics[width=0.85\columnwidth]{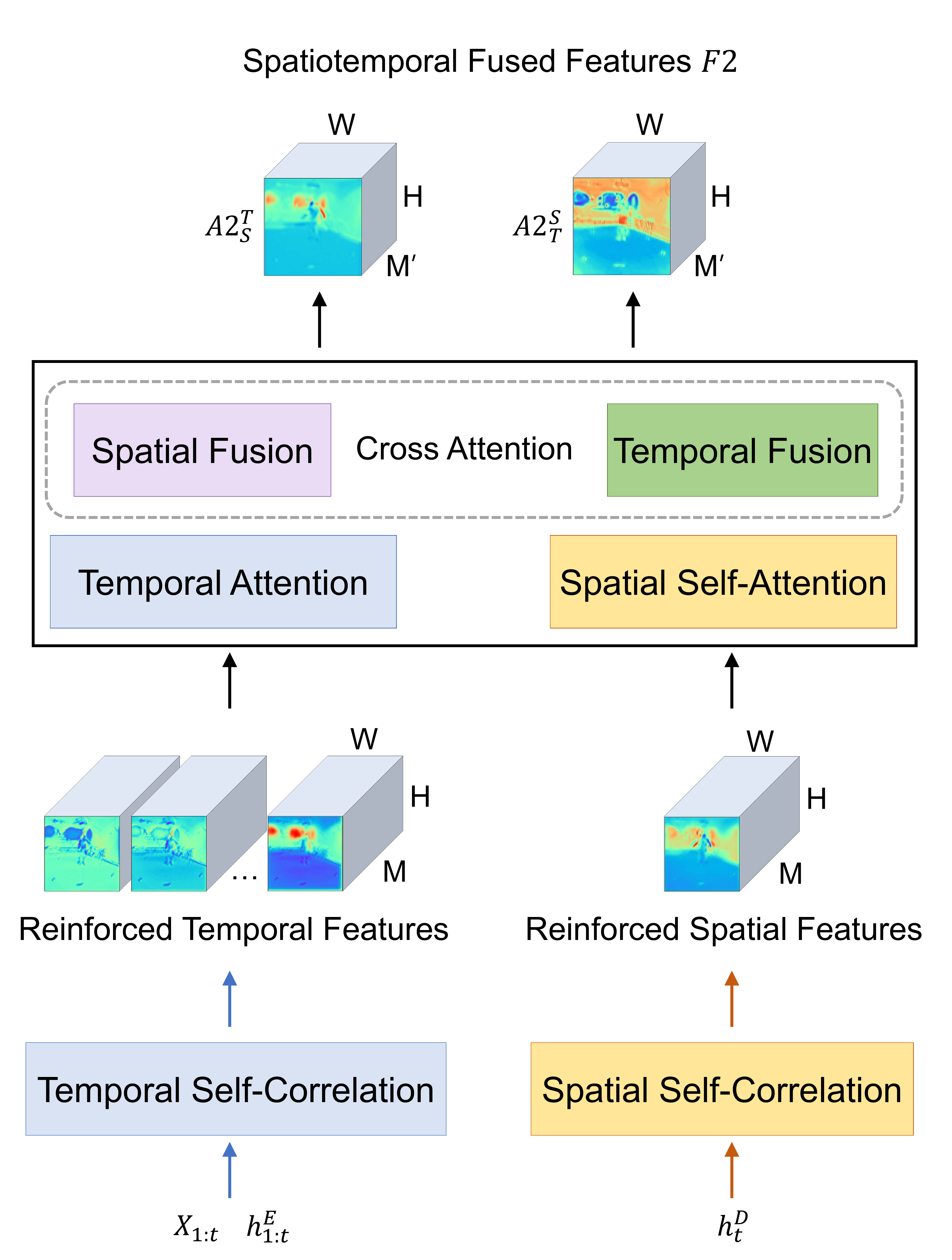}
   \caption{RFA module of SRVP. This module generates reinforced spatiotemporal features and then performs temporal attention, spatial self-attention, and cross-attention.}
   \label{fig4}
\end{figure}

The recurrent memory outputs, $h_{1:t}^{E}$ and $h_{t, 1:L}^{D}$, contain $M$ hidden states per pixel, encoding both temporal correlations between frames and spatial information within frames. Instead of directly utilizing these feature representations, applying contrast enhancement prior to the attention mechanism enables a more pronounced extraction of features related to object shape. Therefore, we designed the RFA module (\cref{fig4}), which performs feature reinforcement and subsequently generates the spatiotemporal fused features based on the attention mechanism used in the SA module.

The main methods for constructing SRVP are as follows:

\noindent \textbf{Temporal Attention}
Our approach employs the conventional scaled dot-product attention mechanism \cite{selfatt}. The target features, $h_{t, 1:L}^{D} \in \mathbb{R}^{L \times MHW}$, are used as the query, whereas the key and value are derived from the reference features, $h_{1:t}^{E} \in \mathbb{R}^{N \times MHW}$. The temporal attention process is then formulated as follows:
\begin{equation}\label{eq2}
    \begin{split}
    &\omega = h_{t, 1:L}^{D}  \otimes {(h_{1:t}^{E})}^{\top} \\
    &\text{A}^T = \text{Softmax}(\frac{\text{Norm}(\omega)}{d})\otimes h_{1:t}^{E}
    \end{split}
\end{equation}
where $\otimes$ denotes matrix multiplication, $d$ is $\sqrt{MHW}$, and $\omega \in \mathbb{R}^{L \times N}$ represents the similarity score between the target and the reference features. The L-wise L2 normalization is applied to $\omega$ before applying the softmax function. Through this process, temporal context features $\text{A}^T \in \mathbb{R}^{M \times HW}$ containing the reference feature information most relevant to the currently predicted image are obtained.

\noindent \textbf{Spatial Self-Attention}
The target features $h_{t, 1:L}^{D}$ are normalized along the $L$-axis and transformed to the shape of $h_{t, n}^{D} \in \mathbb{R}^{M \times HW}$. Thereafter, they are embedded into $Q, K, \text{and } V \in \mathbb{R}^{M \times HW}$ as follows:
\begin{equation}\label{eq3}
    \begin{split}
    &Q = W_q(h_{t, n}^{D}) \\
    &K = W_k(h_{t, n}^{D}) \\
    &V = W_v(h_{t, n}^{D})
    \end{split}
\end{equation}
where $W_q$, $W_k$, and $W_v$ represent channel-wise (M-wise) linear projections. Similar to temporal attention, we apply scaled dot-product attention to derive the spatial context features $\text{A}^S \in \mathbb{R}^{M \times HW}$:
\begin{equation}\label{eq4}
    \begin{split}
    &\text{A}^S = \text{Softmax}(\frac{Q\otimes{K}^{\top}}{d_k})\otimes V
    \end{split}
\end{equation}
where $d_k$ is $\sqrt{HW}$. $\text{A}^S$ encodes information regarding the regions within the image that should be emphasized, thereby enhancing spatial feature extraction.

\noindent \textbf{Cross-Attention}
Through the aforementioned processes, $\text{A}^T$ captures temporal context information, whereas  $\text{A}^S$ retains information on short-term variations. Subsequently, these two distinct feature representations are fused to generate dynamic spatiotemporal features. First, a linear projection is applied to $\text{A}^T$ and $\text{A}^S$, similar to that in \cref{eq3}, to generate the query, key, and value, which are used to compute the spatiotemporally fused features $F1 \in \mathbb{R}^{2M \times HW}$ as follows:
\begin{equation}\label{eq5}
    \begin{split}
    &\text{A1}_{S}^{T} = \text{Softmax}(\frac{Q_T\otimes{K_S}^{\top}}{d_k})\otimes V_S \\
    &\text{A1}_{T}^{S} = \text{Softmax}(\frac{Q_S\otimes{K_T}^{\top}}{d_k})\otimes V_T \\
    &F1 = [\text{A1}_{S}^{T}, \text{A1}_{T}^{S}]
    \end{split}
\end{equation}
where $Q_T, K_T, \text{and } V_T \in \mathbb{R}^{M \times HW}$ are produced by linear projections of $\text{A}^T$, whereas $Q_S, K_S, \text{and } V_S \in \mathbb{R}^{M \times HW}$ orignate from $\text{A}^S$. $\text{A1}_{S}^{T} \in \mathbb{R}^{M \times HW}$ represents a spatially fused temporal context feature and $\text{A1}_{T}^{S} \in \mathbb{R}^{M \times HW}$ denotes a temporally fused spatial context feature.

\noindent \textbf{Feature Reinforcement}
Reinforced features ${h'}_{1:t}^{E} \in \mathbb{R}^{N \times MHW}$ and ${h'}_{t}^{D} \in \mathbb{R}^{M \times HW}$ were generated based on different self-correlation maps \cite{selfcorrmap, selfcorrmap2, selfcorrmap3}. Following self-correlation calculations, the attention process employed in the SA module is used to obtain the spatiotemporally fused features $F2 = [\text{A2}_{S}^{T}, \text{A2}_{T}^{S}] \in \mathbb{R}^{2M' \times HW}$, as shown in \cref{fig4}.

\noindent \textbf{Temporal Self-Correlation}
Reinforced temporal features are generated using both the spatial feature map of historical frames and the reference features. First, convolution is applied to each frame of the historical sequence, producing the feature map $X' \in \mathbb{R}^{N \times M \times H \times W}$ as follows:
\begin{equation}\label{eq6}
    \begin{split}
    &X' = \{\text{ConvNet}(X_i)\}_{i=1}^{t}
    \end{split}
\end{equation}
$\text{ConvNet}$ includes 2D convolution layers, batch normalization, and ReLU. This process extracts spatial features from each frame. Subsequently, time-wise (N-wise) softmax and L2 normalization are employed to identify the most active feature map within the past frames. Based on this, the temporal self-correlation map $\psi^T \in \mathbb{R}^{N \times MHW}$ is generated as follows:
\begin{equation}\label{eq7}
    \begin{split}
    &\psi^T = \text{Norm}(X') \otimes \text{Norm}(X' \otimes \text{Softmax}(X')^{\top})^{\top}
    \end{split}
\end{equation}
Finally, the reinforced temporal features ${h'}_{1:t}^{E}$ are obtained as
\begin{equation}\label{eq8}
    \begin{split}
    &{h'}_{1:t}^{E} = \text{LayerNorm}(h_{1:t}^{E} + \psi^T)
    \end{split}
\end{equation}
This process enhances the local contrast of the reference features, enabling the extraction of richer video dynamics. ${h'}_{1:t}^{E}$ is generated during the initial time-step prediction and remains constant for subsequent predictions.

\noindent \textbf{Spatial Self-Correlation}
This process focuses on enhancing the spatial details of $h_{t, 1:L}^{D}$. To effectively utilize both low and high-level features, the hidden states across all layers are concatenated into a single dimension, transforming them into the shape of $h_{t}^{D} \in \mathbb{R}^{LM \times HW}$. The spatial self-correlation map $\psi^S \in \mathbb{R}^{M \times HW}$ is computed in a similar manner as in \cref{eq7}:
\begin{equation}\label{eq9}
    \begin{split}
    &\psi^S = \text{Norm}(h_{t}^{D}) \otimes \text{Norm}(h_{t}^{D} \otimes \text{Softmax}(h_{t}^{D})^{\top})^{\top}
    \end{split}
\end{equation}
Channel-wise (LM-wise) softmax and L2 normalization are applied in this case. The reinforced spatial features ${h'}_{t}^{D}$ are then generated as follows:
\begin{equation}\label{eq10}
    \begin{split}
    &{h'}_{t}^{D} = \text{ConvNet}(\text{LayerNorm}(h_{t}^{D} + \psi^S))
    \end{split}
\end{equation}
$\text{ConvNet}$ comprises a 2D convolution layer and Tanh activation function, allowing for the extraction of a spatially sharper feature map. Notably, this approach enhances contrast in regions where motion occurs (e.g., a person’s limbs).

\noindent \textbf{Output Layer}
The SA module generates spatiotemporally fused features $F1$ using $h_{1:t}^{E}$ and $h_{t, 1:L}^{D}$, whereas the RFA module utilizes ${h'}_{1:t}^{E}$, and ${h'}_{t}^{D}$ to compute $F2$. Finally, the output layer predicts $X_{t+1}$ as follows:
\begin{equation}\label{eq11}
    \begin{split}
    &X'_{t+1} = \text{ConvNet}(\text{Concat}(h_{t}^{D}, F1, F2))
    \end{split}
\end{equation}
$\text{ConvNet}$ comprises a 2D convolution layer and sigmoid activation function. In the conventional baseline model (\cref{fig1}), predictions are generated using only $h_{t, 1:L}^{D}$. In contrast, SRVP uses enhanced spatiotemporal features, such as $F1$ and $F2$, producing more refined predictions. 

%% file: sec/4_experiments.tex
\section{Experiments}
\label{sec:exp}

\subsection{Benchmarks}
We assessed the performance of our proposed model using three established benchmark datasets: Moving MNIST \cite{vplstm}, KTH Action \cite{kth}, and Human3.6M \cite{h36m}.

Moving MNIST \cite{vplstm} is a widely used benchmark for VP, comprises 64×64 grayscale image sequences. We followed the procedure in \cite{mimo} to generate 10,000 training sequences and used the 10,000 test sequences provided in \cite{vplstm}. The models were trained to predict the 10 subsequent frames after receiving 10 input frames. Additionally, we evaluated the models' ability to capture long-term dependencies by extending the prediction length to 30 frames.

KTH Action \cite{kth} contains grayscale videos showing various human actions, categorized into six groups. Each category includes 25 subjects, performing the actions under four different conditions: outdoors, outdoors with scale variations, outdoors with different attire, and indoors. Consistent with previous research, we used subjects 1– 16 for training and 17–25 for testing. Images were center-cropped and resized to 64×64. All models were trained to predict the next 10 frames by taking 10 observed frames.

Human3.6M \cite{h36m} contains color video sequences depicting 17 different human actions. Subjects S1, S5, S6, S7, and S8 were used for training, whereas S9 and S11 were reserved for testing. Our experiments focused on the Walking and WalkTogether scenarios, with images resized from 1000×1000 to 100×100. The models were trained to predict the next 4 frames by taking 4 observed frames.

\subsection{Reference Models}
We built five RNN-based VP models by equipping the recurrent layers of the baseline architecture with several previously proposed spatiotemporal memory units, including ConvLSTM \cite{convlstm}, ST-LSTM \cite{predrnn}, Causal LSTM \cite{predrnn2}, MIM \cite{mim}, and ConvGRU (\cref{sec:convgru}). For the Causal LSTM configuration, we followed the method outlined in the original study \cite{predrnn2}, incorporating a GHU into the first recurrent layer for five layers. To evaluate the ability of SRVP to mitigate the error accumulation problem inherent in RNN-based models, we compared its performance with that of RNN-free models. We selected MIMO-VP \cite{mimo}, which employs a Transformer-based prediction approach, and SimVP \cite{simvp}, which employs a CNN-based method, as our reference models. To the best of our knowledge, these are the most recently published state-of-the-art models.

\subsection{Implementation Details}
We trained the SRVP and the RNN-based VP models using the binary cross-entropy loss function. The learning rate was selected from the set $\{1e-4, 1e-5, 5e-6\}$, depending on the specific dataset. Model optimization was performed using the RMSProp optimizer and a cosine annealing scheduler. All models were trained in more than 150 epochs, with a batch size of 8. For the RNN-free models, we adhered to the configurations described in their original study. Model performance was evaluated using mean squared error (MSE), peak signal-to-noise ratio (PSNR), and the structural similarity index measure (SSIM). Our implementation is available at \url{https://github.com/yuseonk/SRVP}.

\subsection{Evaluations}
In this section, we conduct both quantitative and qualitative evaluations. All qualitative results are illustrated in the supplementary material.

\noindent \textbf{Moving MNIST}
This dataset comprised sequences featuring two simultaneously moving digits, identical or different. The primary challenge involved accurately predicting the shapes of the digits, particularly when they overlapped or moved apart. 
The quantitative results for a 10-frame prediction horizon are presented in \cref{tab:mnist}. SRVP demonstrated improvements of up to 39.02\% in MSE and 36.36\% in SSIM compared to traditional RNN-based models. However, its performance was lower than that of the RNN-free models.
\begin{table}[h]
    \caption{Performance comparison on the Moving MNIST dataset. Lower MSE and higher SSIM values indicate better results. The results are averaged for all predicted frames.}
    \centering
    \renewcommand{\arraystretch}{1.1} 
    \setlength{\tabcolsep}{2pt}       
    \begin{tabular}{c c c c c}
        \toprule
        \multirow{2}{*}{\textbf{Method}} & 
        \multicolumn{2}{c}{\textbf{10 $\rightarrow$ 10}} & 
        \multicolumn{2}{c}{\textbf{10 $\rightarrow$ 30}} \\
        \cmidrule(lr){2-3} \cmidrule(lr){4-5}
        & \textbf{MSE $\downarrow$} & \textbf{SSIM $\uparrow$ } & \textbf{MSE $\downarrow$} & \textbf{SSIM $\uparrow$ } \\
        \midrule
        ConvLSTM \cite{convlstm}     & 1419.41 & 0.5661 & 1272.60 & 0.4963 \\
        ST-LSTM \cite{predrnn}       & 1533.24 & 0.5481 & 1324.94 & 0.4636 \\
        Causal LSTM \cite{predrnn2}  & 1162.65 & 0.6693 & 1303.48 & 0.4886 \\
        MIM \cite{mim}               & 1385.35 & 0.5950 & 1303.11 & 0.4908 \\
        ConvGRU (\cref{sec:convgru}) & 1033.93 & 0.7082 & 1257.30 & 0.5221 \\
        MIMO-VP \cite{mimo}          & \textbf{401.59} & \textbf{0.9049} & \textbf{1021.55} & \textbf{0.6564} \\
        SimVP \cite{simvp}           & 673.01 & 0.8283 & 1280.05 & 0.5167 \\
        \midrule
        SRVP                         & 935.02 & 0.7474 & 1173.18 & 0.5533 \\
        \bottomrule
    \end{tabular}
    \label{tab:mnist}
\end{table}

\begin{figure}[t]
  \centering
   \includegraphics[width=0.9\columnwidth]{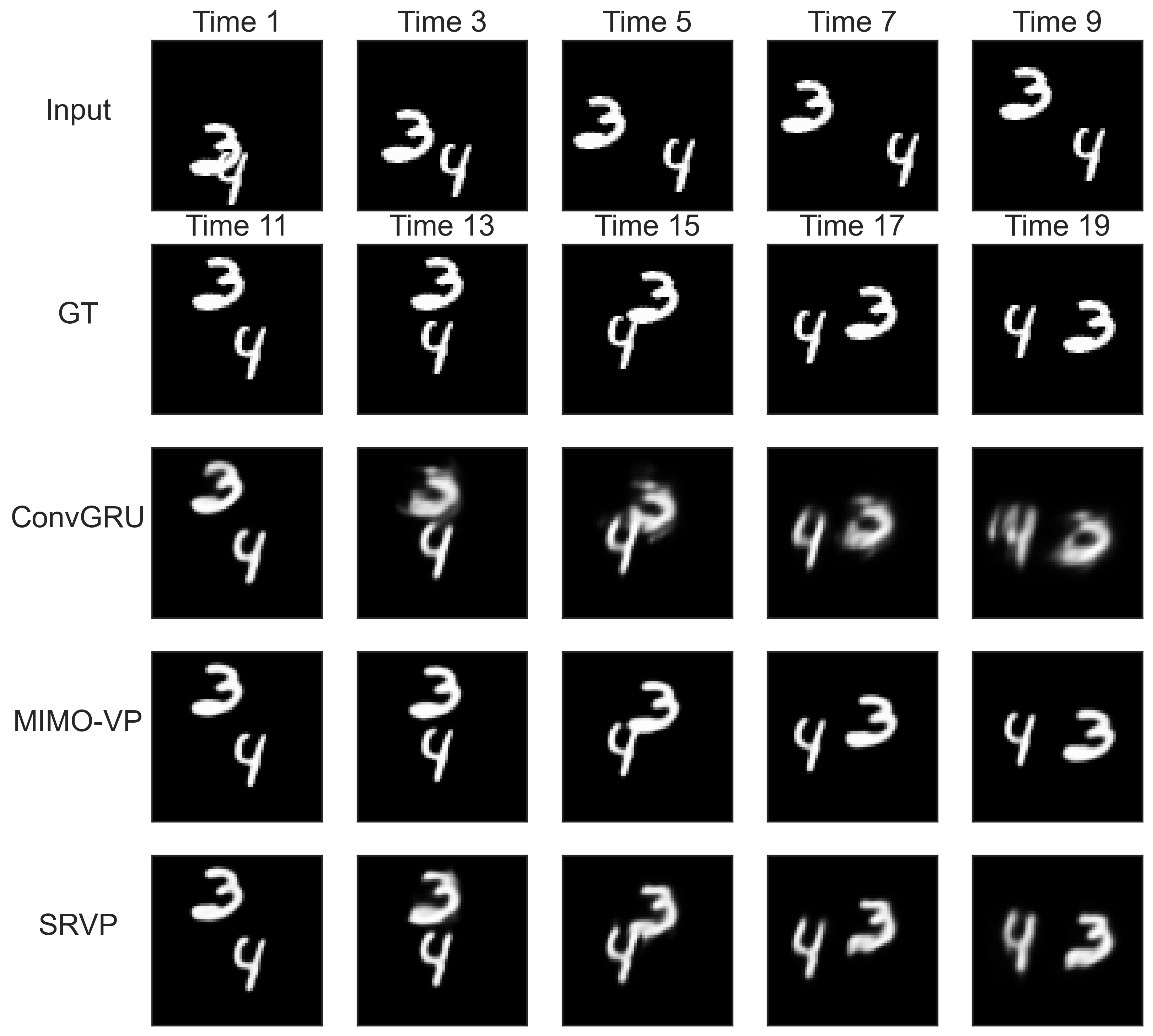}
   \caption{Prediction results for the Moving MNIST (10 $\rightarrow$ 10). GT indicates ground-truth sequences. ConvGRU is the best-performing RNN-based model, and MIMO-VP is the best-performing RNN-free model.}
   \label{result_m10}
\end{figure}

Nonetheless, as illustrated in \cref{result_m10}, SRVP significantly reduced the blurring artifacts observed in the traditional RNN-based models, achieving visual quality comparable to the RNN-free models. By incorporating two attention modules, SA and RFA, into ConvGRU, SRVP introduces a modest increase in computational cost (approximately 6.3\% in FLOPs) compared to ConvGRU. However, this trade-off is justified by the significant improvement in prediction accuracy (approximately 9.6\% in MSE) and enhanced capture of fine-grained motion details. These findings indicate that SRVP more effectively preserves the spatial representation of objects, thereby mitigating the error accumulation problem that affects long-term predictions in RNN-based models.

\begin{figure}[t]
  \centering
   \includegraphics[width=0.95\columnwidth]{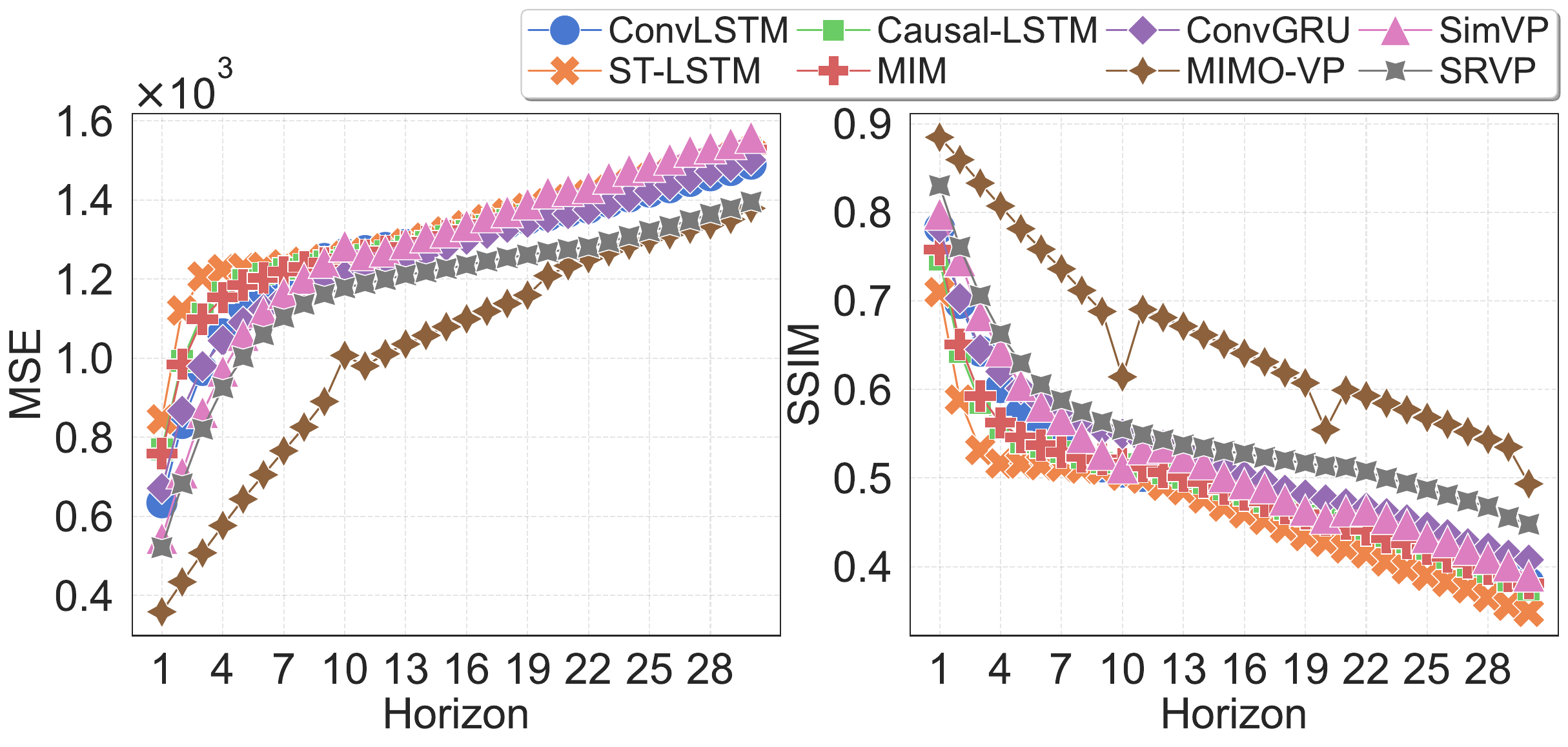}
   \caption{Frame-wise quantitative results for the Moving MNIST dataset (10 $\rightarrow$ 30).}
   \label{metric_m30}
\end{figure}

\begin{figure}[t]
  \centering
   \includegraphics[width=0.8\columnwidth]{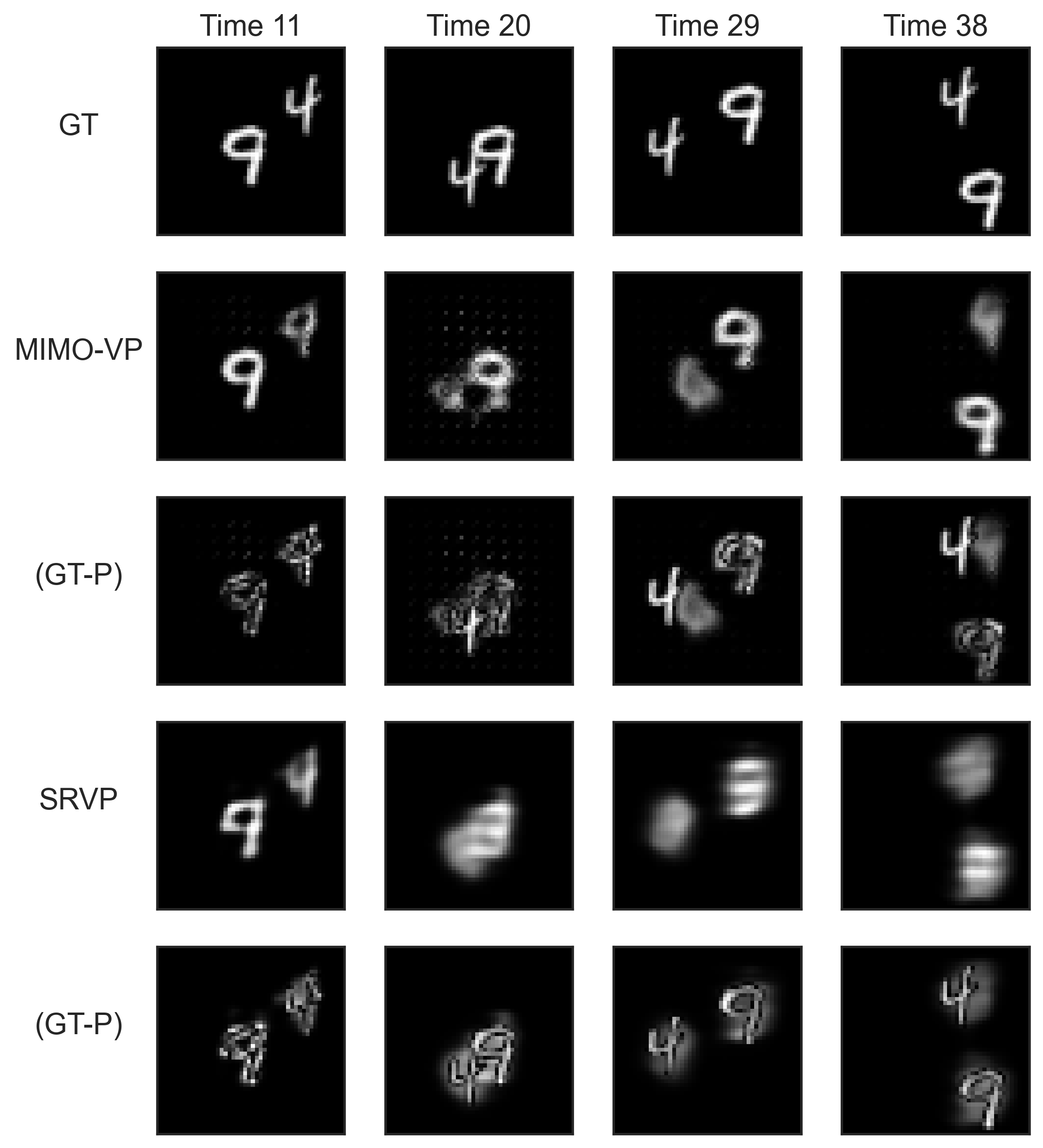}
   \caption{Prediction results for the Moving MNIST dataset (10 $\rightarrow$ 30). GT represents ground-truth sequences, whereas GT-P indicates the difference between the ground-truth and predicted frames. MIMO-VP shows dot artifacts at Time 20.}
   \label{result_m30}
\end{figure}

Furthermore, in the long-term dependency evaluation, SRVP showed improvements of 8.35\% in MSE and 7.08\% in SSIM compared to SimVP. Although MIMO-VP achieved the best quantitative performance, it made inconsistent predictions over time, with large errors manifesting in the middle of sequences (\cref{metric_m30,result_m30}). Therefore, SRVP might be more effective than RNN-free models in capturing long-term temporal relationships.

\noindent \textbf{KTH Action}
The key difficulty with this dataset is estimating human motion over long durations. This requires the model to simultaneously predict changes in position within the scene and the movements of limbs. The quantitative results listed in \cref{tab:kth} reveal that SRVP achieved superior performance on this challenging task. SRVP demonstrated improvements of up to 59.09\% in MSE, 29.69\% in PSNR, and 38.29\% in SSIM compared to reference models.

\begin{figure}[t]
  \centering
   \includegraphics[width=\columnwidth]{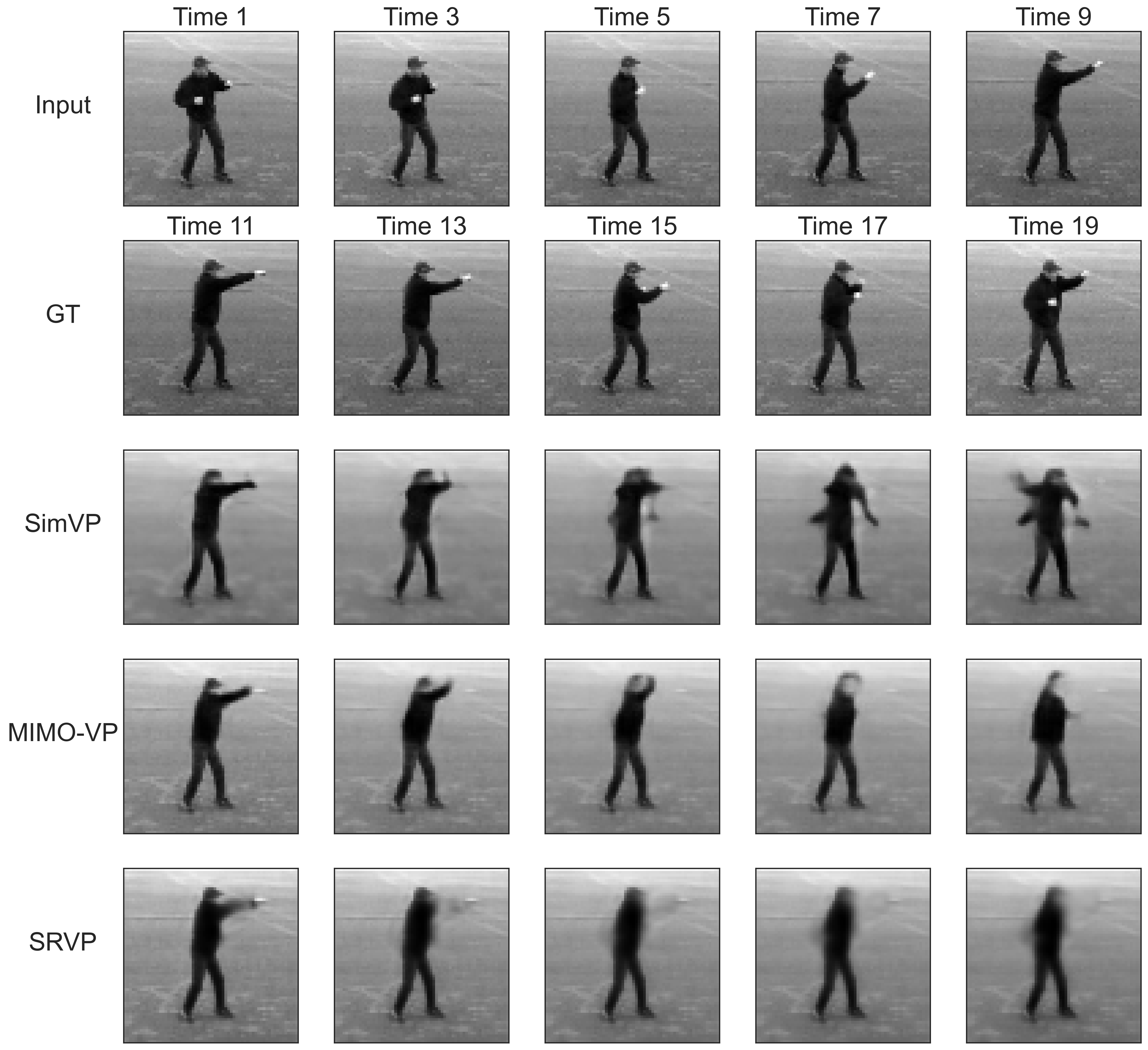}
   \caption{Prediction results for the KTH dataset (10 $\rightarrow$ 10). GT indicates ground-truth sequences.}
   \label{result_kth}
\end{figure}

\cref{result_kth} provides a qualitative comparison between RNN-free models and SRVP. SRVP effectively captures the overall human motion trend, whereas the RNN-free models fail to accurately predict the trajectory. This leads to considerable errors in the position where movement occurs when performing a pixel-by-pixel comparison. Moreover, efforts are still needed to preserve the spatial details. Future improvements should distinguish the high-motion areas from the background.
\begin{table}[t]
  \caption{Performance comparison on the KTH dataset. Lower MSE and higher PSNR and SSIM values indicate better results. The results are averaged for all predicted frames.}
  \centering
  \begin{tabular}{P{3cm}P{1.2cm}P{1.2cm}P{1.2cm}}
    \toprule
    \textbf{Method} & \textbf{MSE $\downarrow$} & \textbf{PSNR $\uparrow$} & \textbf{SSIM $\uparrow$} \\
    \midrule
    ConvLSTM \cite{convlstm}     & 936.69 & 17.1198 & 0.5322 \\
    ST-LSTM \cite{predrnn}       & 386.21 & 21.9665 & 0.7294 \\
    Causal LSTM \cite{predrnn2}  & 408.14 & 21.5750 & 0.7158 \\
    MIM \cite{mim}               & 445.27 & 21.3214 & 0.6947 \\
    ConvGRU (\cref{sec:convgru}) & 418.09 & 21.5636 & 0.7115 \\
    MIMO-VP \cite{mimo}          & 502.15 & 21.0167 & 0.7276 \\
    SimVP \cite{simvp}           & 452.65 & 21.4365 & 0.7179 \\
    \midrule
    SRVP                         & \textbf{383.16} & \textbf{22.0322} & \textbf{0.7360} \\
    \bottomrule
  \end{tabular}
  \label{tab:kth}
\end{table}

\begin{figure}[t]
  \centering
   \includegraphics[width=\columnwidth]{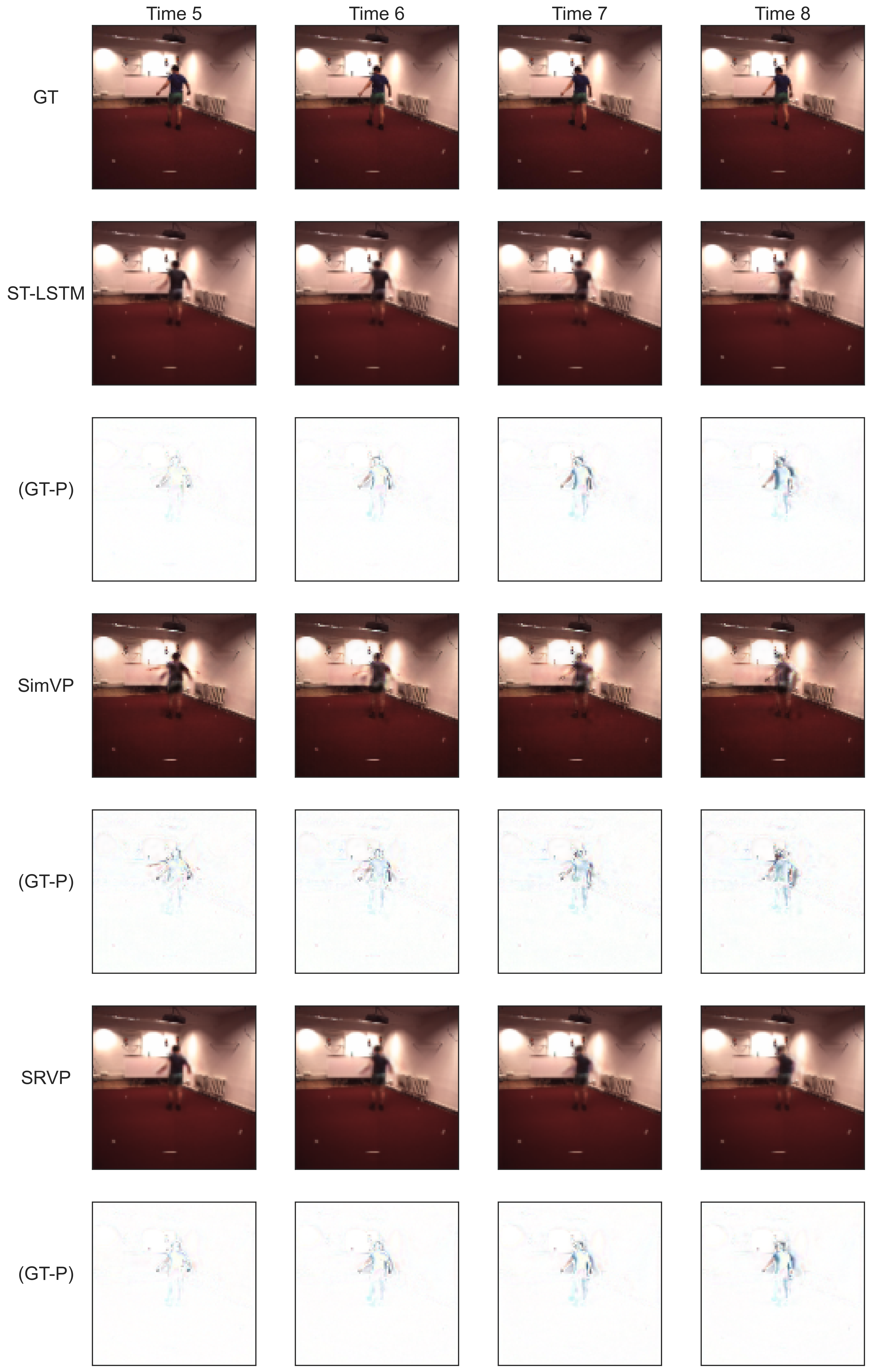}
   \caption{Prediction results for the Human3.6M dataset (4 $\rightarrow$ 4). GT represents the ground-truth sequences, whereas GT-P denotes the difference between the ground-truth and predicted frames.}
   \label{result_h36m}
\end{figure}

\noindent \textbf{Human3.6M}
The three-dimensional nature of the images in this dataset makes it more challenging than the previous two. Quantitative performance was evaluated for each predicted frame (\cref{tab:h36m}). ST-LSTM and SRVP exhibit the highest performance, with a minimal difference of approximately 0.1\% between them. Furthermore, SRVP outperformed RNN-free models by approximately 1.64 to 3.9\%. Although the RNN-free models excelled on less complex datasets such as Moving MNIST, their performance declined as the image size and spatial complexity increased.
\FloatBarrier
\begin{table}[t]
  \caption{Performance comparison for each frame on the Human3.6M dataset. Higher SSIM values indicate better results.}
  \centering
  \renewcommand{\arraystretch}{1.1} 
  \setlength{\tabcolsep}{4pt}       
  \begin{tabular}{c c c c c}
    \toprule
    \multirow{2}{*}{\textbf{Method}} & 
    \multicolumn{4}{c}{\textbf{Horizon}} \\
    \cmidrule(lr){2-5}
    & \textbf{1} & \textbf{2} & \textbf{3} & \textbf{4} \\
    \midrule
    ConvLSTM \cite{convlstm}       & 0.8960 & 0.8799 &  0.8617 & 0.8440 \\
    ST-LSTM \cite{predrnn}          & 0.9656 & \textbf{0.9519} & \textbf{0.9397} & \textbf{0.9285} \\
    Causal LSTM \cite{predrnn2}     & 0.9639 & 0.9502 & 0.9375 & 0.9262 \\
    MIM \cite{mim}                  & 0.9599 & 0.9446 & 0.9313 & 0.9185 \\
    ConvGRU (\cref{sec:convgru})    & 0.9643 & 0.9489 & 0.9354 & 0.9235 \\
    MIMO-VP \cite{mimo}             & 0.9303 & 0.9154 & 0.9034 & 0.8924 \\
    SimVP \cite{simvp}              & 0.9415 & 0.9286 & 0.9194 & 0.9122 \\
    \midrule
    SRVP                            & \textbf{0.9659} & 0.9512 & 0.9386 & 0.9272 \\
    \bottomrule
  \end{tabular}
  \label{tab:h36m}
\end{table}
\cref{result_h36m} presents a qualitative comparison between the best-performing RNN-based model, the best-performing RNN-free model, and SRVP. The frames predicted by SRVP more accurately retain the human shape than those predicted by ST-LSTM, with reduced residual artifacts on the moving arms. In contrast, SimVP exhibits inconsistent image texture and significant errors, not only in the human figure but also in the background.

\noindent \textbf{Ablation Study} We also conducted an ablation study to assess the individual contribution of the proposed attention modules. \cref{tab:ablations} presents the results, indicating that the RFA module is the most critical component of SRVP. This finding suggests that performing attention based on spatiotemporally enhanced contrast features is more effective than directly using hidden state information from recurrent memory.
\begin{table}[t]
  \caption{Ablation results for the Moving MNIST dataset (10 $\rightarrow$ 10). Without-CrossAtt indicates the results obtained after removing the cross-attention process in both SA and RFA.}
  \centering
  \begin{tabular}{P{2.6cm}P{1.1cm}P{1.2cm}P{1.2cm}}
    \toprule
    \textbf{Method} & \textbf{MSE $\downarrow$} & \textbf{PSNR $\uparrow$} & \textbf{SSIM $\uparrow$} \\
    \midrule
    SRVP             & 935.02 & 18.6160 & 0.7474 \\
    Without-SA       & 968.64 & 18.4719 & 0.7365 \\
    Without-RFA      & \textbf{1023.84} & \textbf{18.2027} & \textbf{0.6983} \\
    Without-CrossAtt & 973.16 & 18.4690 & 0.7314 \\
    \bottomrule
  \end{tabular}
  \label{tab:ablations}
\end{table}

%% file: sec/5_conclusions.tex
\section{Conclusions}
This paper presents SRVP, a VP model designed to enhance object representation retention for improved long-term forecasting. SRVP integrates two attention-based modules, SA and RFA, to jointly capture temporal dependencies and spatial correlations. This dual-attention mechanism mitigates the common issue of object detail degradation in long-term VP, where moving objects often blur over time. Empirical evaluations on three benchmark datasets demonstrate that SRVP effectively reduces image degradation in RNN-based models while achieving performance comparable to RNN-free architectures. These results underscore the benefits of our attention mechanisms in improving VP model accuracy and robustness. Future research should explore integrating segmentation techniques to refine pixel-level object representations further.

%% file: sec/ack.tex
\section*{Acknowledgments}
This research was supported by the Institute of Information \& Communications Technology Planning \& Evaluation (IITP, under grant number RS-2024-00397359) and the Korea Institute of Science and Technology Information (KISTI, under grant number K25L1M2C2-01).